# Diffusion of Context and Credit Information in Markovian Models

**Yoshua Bengio**　　　　　　　　　　　　　　　　　　　　　　　　BENGIOY@IRO.UMONTREAL.CA
*Dept. I.R.O., Université de Montréal,*
*Montreal, Qc, Canada H3C-3J7*
**Paolo Frasconi**　　　　　　　　　　　　　　　　　　　　　　　PAOLO@MCCULLOCH.ING.UNIFI.IT
*Dip. di Sistemi e Informatica, Università di Firenze*
*Via di Santa Marta 3 - 50139 Firenze (Italy)*

## Abstract

This paper studies the problem of ergodicity of transition probability matrices in Markovian models, such as hidden Markov models (HMMs), and how it makes very difficult the task of learning to represent *long-term context* for sequential data. This phenomenon hurts the *forward* propagation of long-term context information, as well as *learning* a hidden state representation to represent long-term context, which depends on propagating credit information *backwards* in time. Using results from Markov chain theory, we show that this problem of diffusion of context and credit is reduced when the transition probabilities approach 0 or 1, i.e., the transition probability matrices are sparse and the model essentially deterministic. The results found in this paper apply to learning approaches based on continuous optimization, such as gradient descent and the Baum-Welch algorithm.

## 1. Introduction

Problems of learning on temporal domains can be significantly hindered by the presence of long-term dependencies in the training data. A sequence of random variables (e.g., a sequence of observations $\{\boldsymbol{y}_1, \boldsymbol{y}_2, \ldots \boldsymbol{y}_t, \ldots \boldsymbol{y}_T\}$, denoted $\boldsymbol{y}_1^T$) is said to exhibit long-term dependencies if the variables $\boldsymbol{y}_t$ at a given time $t$ are significantly dependent on the variables $\boldsymbol{y}_{t_0}$ at much earlier times $t_0 \ll t$. In these cases, a system trained on this data (e.g., to model its distribution, or make classifications or predictions) has to be able to store for arbitrarily long durations bits of information in its state variable, called $x_t$ here. In general, the difficulty is not only to represent these long-term dependencies, but also to *learn* a representation of past *context* which takes them into account. *Recurrent neural networks* (Rumelhart, Hinton, & Williams, 1986; Williams & Zipser, 1989), for example, have an internal state and a rich expressive power that provide them with the necessary long-term memory capabilities.

Algorithms that could efficiently learn to represent long-term context would be useful in many areas of Artificial Intelligence. For example, they could be applied to many problems in natural language processing, both at the symbolic level (e.g., learning grammars and language models), and subsymbolic level (e.g., modeling prosody for speech recognition or synthesis).

In order to train the learning system, however, an effective mechanism of credit assignment through time is needed. To change the parameters of the system in order to change the internal state of the system at time $t$, so as to "improve" the internal state of the system later in the sequence, one can recursively propagate credit or error information backwards in time. For example, the Baum-Welch algorithm for HMMs (Baum, Petrie, Soules, & Weiss, 1970; Levinson, Rabiner, & Sondhi, 1983) and the back-propagation through time algorithm for recurrent neural networks (Rumelhart et al., 1986) rely on such kind of recursion. Numerous gradient-descent based algorithms have been proposed for solving the credit assignment problems in recurrent networks (e.g., Rumelhart et al., 1986; Williams & Zipser, 1989). Yet, many researchers have found practical difficulties in training recurrent networks to perform tasks in which the temporal contingencies present in the input/output sequences span long intervals (Bengio, Simard, & Frasconi, 1994; Mozer, 1992; Rohwer, 1994). Bengio et al. (1994) have also found theoretical reasons for this difficulty and proved a negative result for parametric dynamical systems with a non-linear state to next-state recurrence[1] $x_t = f_t(x_{t-1})$: it will be increasingly difficult to train such as system with gradient descent as the duration of the dependencies to be captured increases. Let $J$ be the matrix of partial derivatives of the state to next-state function, $J_{ij} = \frac{\partial x_{t,i}}{\partial x_{t-1,j}}$. A mathematical analysis of the problem shows that, depending on the norm $|J|$ of the Jacobian matrix $J$, one of two conditions arises in such systems. When $|J| < 1$, the dynamics of the network allow it to reliably store bits of information for arbitrary durations, even with bounded input noise; however, gradients with respect to an error at a given time step vanish exponentially fast as one propagates them backward in time. On the other hand, when $|J| > 1$, gradients can flow backward, but the system is locally unstable and cannot reliably store bits of information for a long time. Bengio et al. (1994) showed how this hurts the learning of long-term dependencies by putting exponentially more weight on the influence of short-term dependencies (in comparison to long-term dependencies) over the gradient of a cost function with respect to trainable parameters. The above negative result applies to non-linear parameterized dynamical systems such as most recurrent networks, but not to linear probabilistic models such as hidden Markov models (HMMs). These models are a special case of our previous result in which the $\infty$-norm $|J| = 1$, because this matrix is a *stochastic* matrix, i.e., a matrix $A$ of transition probabilities $A_{ij} = P(x_t = j | x_{t-1} = i)$, where the state variable $x_t$ can take a finite number of values.

The main contribution of this paper is therefore an extension of the negative results found by Bengio et al. (1994) to the case of *Markovian models*, which include standard HMMs (Baum et al., 1970; Levinson et al., 1983) as well as variations of HMMs such as Input/Output HMMs (IOHMMs) (Bengio & Frasconi, 1995b), and Partially Observable Markov Decision Processes (POMDPs) (Sondik, 1973, 1978; Chrisman, 1992). We find that in general, a phenomenon of diffusion of context and credit assignment, due to the ergodicity of the transition probability matrices, hampers both the representation and the learning of long-term context in the hidden state variable.

Both homogeneous and non-homogeneous Markovian models are considered. *Homogeneous* here means that the transition probabilities of the Markov model are constant over time $t$. *Non-homogeneous* means that these transition probabilities are allowed to be different for each time step, e.g., as a function of an external input that may be different at each time step. In the homogeneous case (e.g., standard HMMs), such models can learn the distribution $P(\boldsymbol{y}_1^T)$ of output sequences $\boldsymbol{y}_1^T = \boldsymbol{y}_1, \boldsymbol{y}_2, \ldots, \boldsymbol{y}_T$ by associating an output distri-

---

1. For example, in the case of a recurrent neural network with recurrent weight matrix $W$ and input vector $u_t$ at time $t$, the next-state recurrence is $f_t(x_{t-1}) = \tanh(W x_{t-1} + u_t)$ bution $P(\boldsymbol{y}_t|x_t = i)$ to each value $i$ of the discrete state variable $x_t$. In the non-homogeneous case, transition and output distributions are conditional on an input sequence, allowing to model relationships between input and output sequences. In the case of IOHMMs (Bengio & Frasconi, 1995b), one thus learns a model $P(\boldsymbol{y}_1^T|\boldsymbol{u}_1^T)$ of the conditional distribution of an output sequence $\boldsymbol{y}_1^T$ when an input sequence $\boldsymbol{u}_1^T$ is given. This can be used to perform sequence regression or classification, as with recurrent networks. In the case of POMDPs (Sondik, 1973, 1978; Chrisman, 1992), used to control a process with a hidden state, one wants not only to build such a model, but also to select a proper sequence $a_1^T$ of (discrete) actions in order to maximize a discounted sum of future rewards that depends on the action taken, the observed output sequence $\boldsymbol{y}_1^T$ and the estimated distribution of the state trajectory. Note that the sequence of actions $a_1^T$ in POMDPs and the sequence of inputs $\boldsymbol{u}_1^T$ in IOHMMs play a similar role in this paper, inasmuch as both are responsible for the non-homogeneity of the Markov chain. In the following, we shall use the same symbol $\boldsymbol{u}_1^T$ to denote the sequence that controls transition probabilities, i.e. inputs for IOHMMs and actions for POMDPs.

The negative results presented in this paper are directly applicable to learning algorithms such as the EM algorithm (Dempster, Laird, & Rubin, 1977) or other gradient-based optimization algorithms, which rely on gradually and iteratively modifying continuous-valued parameters (such as transition probabilities, or parameters of a function computing these probabilities) in order to optimize a learning criterion.

## 2. Mathematical Preliminaries

A first-order Markovian model is defined by a discrete set of states $\{1, \ldots n\}$, a probabilistic transition function (state to next-state), and a probabilistic output function (state to output). The discrete state variable $x_t$ can take values in $\{1, \ldots n\}$ at each time step. We will write $A_{ij}$ for the element $(i,j)$ of a matrix $A$, $A^n = AA \ldots A$ for the $n^{\text{th}}$ power of $A$, and $(A^n)_{ij}$ for the element $(i,j)$ of $A^n$. See (Rabiner, 1989) for an introduction to HMMs, and (Seneta, 1981) for a basic reference on positive matrices.

The *Markovian* independence assumption implies that the state variable $x_t$ summarizes the past of the sequence: $P(x_t|x_1, x_2, \ldots, x_{t-1}) = P(x_t|x_{t-1})$. Another independence assumption, when the state $x_t$ is hidden but an output $\boldsymbol{y}_t$ is observed, is that the distribution of $\boldsymbol{y}_t$ at time $t$ does not depend on the other past variables when $x_t$ is given. State transitions at time $t$ may depend on the $\boldsymbol{u}_t$ (the current input for IOHMMs or the current action for POMDPs) and can be collected into an $n$ by $n$ transition matrix $A_t$ defined by

$$A_{ij}(\boldsymbol{u}_t) = \mathrm{P}(x_t = j \mid x_{t-1} = i, \boldsymbol{u}_t; \boldsymbol{\theta})$$

where $\boldsymbol{\theta}$ is a vector of adjustable parameters. In the homogeneous case, the transition matrix is constant, i.e., $A_t = A$. The parameters $\boldsymbol{\theta}$ are then usually directly identified with the elements of the transition matrix $A$.

Output emissions $\boldsymbol{y}_t$ depend on $\boldsymbol{u}_t$ and the present state, as specified by the output (also called emission) distribution $P(\boldsymbol{y}_t \mid x_t, \boldsymbol{u}_t; \boldsymbol{\vartheta})$, with parameters $\boldsymbol{\vartheta}$. For example, if the Markov chain is homogeneous and the output values belong to a finite alphabet of cardinality $k$, then the parameters $\boldsymbol{\vartheta}$ can be collected in a $k$ by $n$ matrix $B$, $B_{li} = P(\boldsymbol{y}_t = l \mid x_t = i)$.

An output sequence $\boldsymbol{y}_1^T$ can be generated according to the distribution $P(\boldsymbol{y}_1^T|\boldsymbol{u}_1^T)$ (non-homogeneous case) or $P(\boldsymbol{y}_1^T)$ (homogeneous case) represented by the model, as follows. First an initial state $x_0$ is selected according to a distribution $P(x_0)$ on initial states (usually multinomial, sometimes requiring $n-1$ extra parameters, or a fixed choice of a single initial state). Then the state $x_t$ can be recursively picked in function of the previous state $x_{t-1}$, by choosing an $x_t \in \{1, \ldots, n\}$ according to the multinomial distribution $P(x_t|x_{t-1}, \boldsymbol{u}_t; \boldsymbol{\theta})$. At each time step, an output can then be generated according to the distribution $P(\boldsymbol{y}_t \mid x_t, \boldsymbol{u}_t; \boldsymbol{\vartheta})$.

State transitions can be constrained by a *directed graph* $\mathcal{G}$, whose nodes are associated to the states of the Markov chain. In particular, the probability $P(x_t = i \mid x_{t-1} = j)$ will be constrained to be zero if there is no edge from node $j$ to node $i$.

## 2.1 Learning in Markovian Models

The learning objective is often to maximize the output likelihood $P(\boldsymbol{y}_1^T; \boldsymbol{\Theta})$, or the output likelihood given the input $P(\boldsymbol{y}_1^T \mid \boldsymbol{u}_1^T; \boldsymbol{\Theta})$, where $\boldsymbol{\Theta}$ comprises all the parameters of the model. This can be accomplished with an EM algorithm when the form of the output and transition probability models are simple enough, e.g. in the case of HMMs (Baum et al., 1970; Levinson et al., 1983; Rabiner, 1989) or IOHMMs (Bengio & Frasconi, 1995b). Alternatives, for maximizing the output likelihood or other criteria (such as the more discriminant mutual information between the output sequence and the correct model, Bahl et al. 1986), are usually based on some gradient-based optimization algorithm, requiring the computation of the gradient of the learning criterion with respect to the model parameters. In all of these cases, the learning algorithms perform products involving the transition probability matrices (Bengio & Frasconi, 1995a, 1995b), such as

$$\begin{aligned}\alpha_{i,t} &= \mathrm{P}(\boldsymbol{y}_1^t, x_t = i \mid \boldsymbol{u}_1^t) = P(\boldsymbol{y}_t \mid x_t = i, \boldsymbol{u}_t) \sum_\ell A_{\ell i}(\boldsymbol{u}_t)\alpha_{\ell,t-1}\\ \beta_{i,t} &= \mathrm{P}(\boldsymbol{y}_t^T \mid x_t = i, \boldsymbol{u}_t^T) = \sum_\ell A_{i\ell}(\boldsymbol{u}_{t+1})P(\boldsymbol{y}_t \mid x_{t+1} = l, \boldsymbol{u}_{t+1})\beta_{\ell,t+1}.\end{aligned} \quad (1)$$

where the overall output likelihood is obtained from the final time step:

$$P(\boldsymbol{y}_1^T \mid \boldsymbol{u}_1^T) = \sum_i \alpha_{i,T}.$$

Note that if $L$ is the learning criterion and $\beta_{i,T} = \frac{\partial L}{\partial \alpha_{i,T}}$, then $\beta_{i,t} = \frac{\partial L}{\partial \alpha_{i,t}}$. In terms of matrices, we can write

$$\begin{aligned}\boldsymbol{\alpha}_t &= \Lambda_t A_t' \cdots \Lambda_1 A_1' \boldsymbol{\alpha}_0\\ \boldsymbol{\beta}_t &= A_t \Lambda_t \cdots A_T \Lambda_T \boldsymbol{\beta}_T\end{aligned} \quad (2)$$

where $\boldsymbol{\alpha}_t = [\alpha_{1,t} \ldots \alpha_{n,t}]'$, $\boldsymbol{\beta}_t = [\beta_{1,t} \ldots \beta_{n,t}]'$ and $\Lambda_t$ is a diagonal matrix of emission probabilities $P(\boldsymbol{y}_t|x_t = i, \boldsymbol{u}_t)$ (for the i[th] element). The matrix $A_t$ contains the transition probabilities at time $t$, i.e. $(A_t)_{ij} = \mathrm{P}(x_t = j \mid x_{t-1} = i, \boldsymbol{u}_t; \boldsymbol{\theta})$. It can be easily verified that the compact notation

$$A^{(t_0, t)} = A_{t_0} A_{t_0+1} \cdots A_{t-1} A_t \quad (3)$$

for products of matrices[2] can be used to describe the effect of the distribution of the state $x_{t_0}$ at time $t_0$ on the distribution of the state $x_t$ at time $t > t_0$: $A^{(t_0,t)}_{ij} = \mathrm{P}(x_t = j \mid x_{t_0} = i, \boldsymbol{u}_{t_0}^t; \boldsymbol{\theta})$.

---

2. To verify equation (3), just apply recursively the simple decomposition rule of probabilities $\mathrm{P}(a) = \sum_b \mathrm{P}(a \mid b)\mathrm{P}(b)$.

Therefore, we will study how this product evolves under various conditions, when $t - t_0$ increases (for long-term dependencies). We will find in what (rather general) conditions $A^{(t_0,t)}$ tends to become ill-conditioned, more precisely, when $x_t$ becomes more and more independent of $x_{t_0}$ as $t - t_0$ increases. In Section 4.2, we also discuss equations (2) as $T - t$ increases. In the following subsection we first introduce some standard mathematical tools for studying such products of non-negative matrices.

## 2.2 Definitions

**Definition 1** *(Non-negative matrices) A matrix $A$ is said to be* non-negative, *written $A \geq 0$, if $A_{ij} \geq 0 \;\; \forall i,j$.*

*Positive* matrices are defined similarly.
By extension, we will also write $A \geq B$ when $\forall i, j$, $A_{ij} \geq B_{ij}$.

**Definition 2** *(Stochastic matrices) A non-negative square matrix $A \in \mathbf{R}^{n \times n}$ is called* row stochastic *(or simply* stochastic *in this paper) if $\sum_{j=1}^{n} A_{ij} = 1 \;\; \forall i = 1 \ldots n$.*

**Definition 3** *(Allowable matrices) A non-negative matrix is said to be* row [column] allowable *if every row [column] sum is positive. An* allowable *matrix is both row and column allowable.*

A non-negative matrix can be associated to the directed transition graph $\mathcal{G}$ that constrains the Markov chain. The *incidence matrix* $\tilde{A}$ corresponding to a given non-negative matrix $A$ is the 0-1 matrix obtained by replacing all positive entries of $A$ by a 1. The incidence matrix of $A$ is a connectivity matrix corresponding to the graph $\mathcal{G}$ (assumed to be connected here). Some algebraic properties of $A$ are described in terms of the *topology* of $\mathcal{G}$. Indices of the matrix $A$ correspond to *nodes* of $\mathcal{G}$ (we will also use "states of the model", talking about a Markovian model).

**Definition 4** *(Irreducible Matrices) A non-negative $n \times n$ matrix $A$ is said to be* irreducible *if for every pair $i, j$ of indices, $\exists \; m = m(i,j)$ positive integer s.t. $(A^m)_{ij} > 0$.*

A matrix $A$ is irreducible if and only if the associated graph is *strongly connected* (i.e., there exists a path between any pair of states $i, j$). A *reducible* matrix is one that is not irreducible. If $\exists k$ s.t. $(A^k)_{ii} > 0$ (i.e., there is a path of length $k$ from node $i$ to itself), $d(i)$ is called the *period* of index $i$ if $d(i)$ is the greatest common divisor (g.c.d.) of those $k$ for which $(A^k)_{ii} > 0$ (i.e., there are also paths of length $k$, $2k$, $3k$, etc..., with $k = d(i)$). In an irreducible matrix all the indices have the same period $d$, which is called the *period* of the matrix. The period of a matrix is the g.c.d. of the lengths of all cycles in the associated transition graph $\mathcal{G}$.

An example of a periodic matrix of period 3 is illustrated by the graph $\mathcal{G}_1$ of Figure 2. All the paths starting from one of the states and returning to it are of length $3k$ for some positive integer $k$.

**Definition 5** *(Primitive matrix) A non-negative matrix $A$ is said to be* primitive *if there exists a positive integer $k$ s.t. $A^k > 0$.*

Therefore, in a graph with a corresponding primitive matrix, one can always find a path of length greater than some $k$ between any two nodes, and if there exists a path of length $k$ between nodes $i$ and $j$, there are also paths of length $k+1$, $k+2$, etc... In the analysis below, we will consider submatrices (and corresponding subgraphs) which are primitive. Note that an irreducible matrix is either periodic or primitive (i.e., of period 1), and that a primitive stochastic matrix is necessarily allowable.

### 2.3 The Perron-Frobenius Theorem

Right eigenvectors $v$ of a matrix $A$ and their corresponding eigenvalues $\lambda$ have the following properties (see Bellman, 1974, for more on eigenvalues and eigenvectors):

$$\text{determinant}(A - \lambda I) = 0.$$

where $I$ is the identity matrix, and

$$Av = \lambda v$$

i.e.,

$$\sum_j A_{ij} v_j = \lambda v_i.$$

Note that for a stochastic matrix $A$ the largest eigenvalue has norm 1, which can be shown as follows. Letting $i = \text{argmax}_j |v_j|$, we obtain

$$|\lambda| = \left|\sum_j A_{ij} \frac{v_j}{v_i}\right| \leq \sum_j |A_{ij}| \frac{|v_j|}{|v_i|} \leq \sum_j A_{ij} \leq 1.$$

Hence all the eigenvalues have norm less or equal to 1. Let us define the vector of ones $\mathbf{1} = [1, 1, \cdots, 1]'$, where $v'$ denotes the transpose of $v$. Since $A\mathbf{1} = \mathbf{1}$ by definition of stochastic matrices, 1 is an eigenvalue and $\mathbf{1}$ is its corresponding right eigenvector.

The following theorem will be useful in characterizing homogeneous products of stochastic matrices (as in HMMs).

**Theorem 1** (Perron-Frobenius Theorem) *Suppose $A$ is an $n \times n$ non-negative primitive matrix. Then there exists an eigenvalue $r$ such that:*

1. *$r$ is real and positive;*

2. *$r$ can be associated with strictly positive left and right eigenvectors;*

3. *$r > |\lambda|$ for any eigenvalue $\lambda \neq r$;*

4. *the eigenvectors associated with $r$ are unique to constant multiples.*

5. *If $0 \leq B \leq A$ and $\beta$ is an eigenvalue of $B$, then $|\beta| \leq r$. Moreover, $|\beta| = r$ implies $B = A$.*

6. *$r$ is a simple root of the characteristic equation $\text{determinant}(A - rI) = 0$.*

[See proof in the book by Seneta, 1981, Theorem 1.1.]

A direct consequence of the Perron-Frobenius theorem for stochastic matrices is therefore the following:

**Corollary 1** *Suppose A is a primitive stochastic matrix. Then its largest eigenvalue is 1 and there is only one corresponding right eigenvector* $\mathbf{1} = [1, 1, \cdots, 1]'$. *Furthermore, all other eigenvalues are less than 1 in modulus.*

**Proof.** $A\mathbf{1} = \mathbf{1}$ by definition of stochastic matrices. As shown above, all the eigenvalues have a modulus less or equal to 1. Thus, we deduce from the Perron-Frobenius Theorem that 1 is the largest eigenvalue, $\mathbf{1}$ is the unique associated eigenvector, and all other eigenvalues $< 1$. □

In the next section we will discuss the consequences of this corollary for HMMs. As shown by Seneta (1981), we should also note that if $A$ is stochastic but *periodic* with period $d$, then $A$ has $d$ eigenvalues of modulus 1 which are the $d$ complex roots of 1.

## 3. Ergodicity

In this section we analyze the case of a primitive transition matrix as well as the general case with a so-called *canonical* re-ordering of the matrix indices (defined below). We introduce *ergodicity coefficients* in order to measure the difficulty in learning long-term dependencies.

### 3.1 Simplest Case: Homogeneous and Primitive

A straightforward application of the Perron-Frobenius theorem and the associated corollary 1 is given in the following theorem.

**Theorem 2** *If $A$ is a primitive stochastic matrix, then as $t \to \infty$, $A^t \to \mathbf{1}v'$ where $v'$ is called the unique stationary distribution of the Markov chain. The rate of approach is geometric.*
[See proof in the book by Seneta, 1981, Theorem 4.2.]

The intuition behind the proof simply relies on the fact that when a matrix $A$ is taken to a certain power $A^n$, it is equivalent to take its eigenvalues to the same power. As we have seen earlier, all the eigenvalues are less or equal to one in modulus. Therefore, the eigenvalues of $A$ which are less than 1 are associated to near zero eigenvalues of $A^n$, as $n \to \infty$. The only eigenvalues which do not converge to zero are those whose modulus is 1. There is only one such eigenvalue in the case of primitive stochastic matrix (associated to the eigenvector $\mathbf{1}$). In the case of periodic matrices of period $d$, discussed below, there are complex eigenvalues whose modulus is 1 and which are among the $d^{\text{th}}$ roots of unity.

We recall that the rank of a matrix $A$ is the dimension of the linear subspace spanned by the eigenvectors of $A$ and corresponds to the number of linearly independent rows (or columns). Since the matrix $A$ obtained by the product $\mathbf{1}v'$ of two vectors has rank 1, we obtain the following from Theorem 2. If $A$ is primitive, then $\lim_{t \to \infty} A^t$ converges to a matrix whose eigenvalues are all 0 except for one eigenvalue $\lambda = 1$ (with corresponding eigenvector $\mathbf{1}$), i.e., the rank of this product converges to 1, which means that its rows are proportional. For a stochastic matrix, row proportionality is equivalent to row equality. Since $(A^{t-t_0})_{ij} = P(x_t = j | x_{t_0} = i)$ it follows that the distribution over the states at time $t > t_0$ becomes gradually independent of the distribution $P(x_{t_0})$ over the states at time $t_0$ as $t - t_0$ increases. This is illustrated in Figure 6, which shows products of 1, 2, 3 and 4 random primitive stochastic matrices, and rapid convergence to row equality, i.e., $P(x_t = j | x_{t_0} = i)$ does not depend any more on $i$ as $t - t_0$ becomes large. It means that, as one moves forward in time, *context information is diffused*, and gradually lost. A consequence of Theorem 2 is therefore that it is very difficult to model long-term dependencies in sequential data using a homogeneous HMM with a primitive transition matrix. After having introduced ergodicity coefficients in the next sections, we will be able to discuss the more general case of non-homogeneous models (such as IOHMMs and POMDPs), as well as, comment on the diffusion of context information in the forward and backward HMM equations (2).

### 3.2 Coefficients of ergodicity

To study products of non-negative matrices and the loss of information about initial state in Markov chains (particularly in the non-homogeneous case), we will define two *coefficients of ergodicity*. First, we introduce the projective distance between vectors $v$ and $w$:

$$d(v', w') = \max_{i,j} \ln\left(\frac{v_i w_j}{v_j w_i}\right).$$

Note that some form of *contraction* takes place when $d(v'A, w'A) \leq d(v', w')$ (Seneta, 1981), i.e., applying the linear operator $A$ to the vectors $v'$ and $w'$ brings them "closer" (according to the above projective distance).

**Definition 6** Birkhoff's contraction coefficient $\tau_B(A)$, *for a non-negative column-allowable matrix $A$, is defined in terms of the projective distance:*

$$\tau_B(A) = \sup_{v,w > 0; v \neq \lambda w} \frac{d(v'A, w'A)}{d(v', w')}.$$

Dobrushin's coefficient $\tau_1(A)$, *for a stochastic matrix $A$, is defined as follows:*

$$\tau_1(A) = \frac{1}{2} \max_{i,j} \sum_k |a_{ik} - a_{jk}|. \tag{4}$$

Both $\tau_B$ and $\tau_1$ are called *proper ergodicity coefficients*, i.e., they have the properties that, firstly, $0 \leq \tau(A) \leq 1$, and secondly, that $\tau(A) = 0$ if and only if $A$ has identical rows (and therefore rank 1). The coefficients of ergodicity quantify the *ergodicity* of a matrix, i.e., at what rate a power of the matrix converges to rank 1. Furthermore, $\tau(A_1 A_2) \leq \tau(A_1)\tau(A_2)$ (Seneta, 1981). Therefore, as discussed in the next section, these coefficients can also be applied to quantify how fast a product of matrices converges to rank 1.

### 3.3 Products of Stochastic Matrices

Let $A^{(1,t)}$ denote a *forward product* of stochastic matrices $A_1, A_2, \cdots A_t$. From the properties of $\tau_B$ and $\tau_1$, if $\tau(A_t) < 1, \forall t > 0$ then $\lim_{t \to \infty} \tau(A^{(1,t)}) = 0$, i.e., $\lim_{t \to \infty} A^{(1,t)}$ has rank 1 and identical rows. Weak ergodicity of a product of matrices is then defined in terms of a proper ergodic coefficient $\tau$ (such as $\tau_B$ or $\tau_1$) converging to 0:

**Definition 7** *(Weak Ergodicity)* *The products of stochastic matrices $A^{(t_0,t)}$ are weakly ergodic if and only if for all $t_0 \geq 0$ as $t \to \infty$, $\tau(A^{(t_0,t)}) \to 0$.*

The following theorem relates weak ergodicity to rank lossage in products of stochastic matrices and, therefore to the problem of learning and representing long-term context.

**Theorem 3** *Let $A^{(1,t)}$ be forward products of non-negative and allowable matrices; then $A^{(1,t)}$ is weakly ergodic if and only if the following conditions both hold:*

1. *$\exists\, t_0$ s.t. $A^{(t_0,t)} > 0\ \forall\, t \geq t_0$;*

2. *$\dfrac{A^{(t_0,t)}_{ik}}{A^{(t_0,t)}_{jk}} \to W_{ij}(t) > 0$ as $t \to \infty$, i.e., rows of $A^{(t_0,t)}$ tend to proportionality.*

[See the proof in the book by Seneta (1981), Lemma 3.3 and 3.4.]

For stochastic matrices, row-proportionality (2nd condition above) is equivalent to row-equality since rows sum to 1. Note that the limit $\lim_{t \to \infty} A^{(t_0,t)}$ itself does not need to exist in order to have weak ergodicity. If such a limit exists and it is a matrix with all rows equal, then the product is said to be *strongly ergodic*.

### 3.4 Canonical Decomposition and Periodic Graphs

Any non-negative matrix $A$ can be rewritten by relabeling its indices in the following *canonical decomposition* (Seneta, 1981), with diagonal blocks $B_i$, $C_i$ and $Q$:

$$A = \begin{pmatrix} B_1 & 0 & \cdots & 0 & \cdots & 0 \\ 0 & B_2 & \cdots & 0 & \cdots & 0 \\ \vdots & \vdots & & & & \vdots \\ 0 & \cdots & C_{s+1} & 0 & \cdots & 0 \\ \vdots & \vdots & & & & \vdots \\ 0 & 0 & \cdots & \cdots & C_r & 0 \\ L_1 & L_2 & \cdots & \cdots & L_r & Q \end{pmatrix} \begin{array}{l} \left.\rule{0pt}{3ex}\right\} \text{Primitive diagonal blocks } B_1,\ldots,B_s \\ \\ \left.\rule{0pt}{3ex}\right\} \text{Periodic diagonal blocks } C_{s+1},\ldots,C_r \end{array} \qquad (5)$$

where the $B_i$ and $C_i$ blocks are irreducible, the $B_i$ blocks are primitive and the $C_i$ blocks are periodic. Define the corresponding sets of states as $S_{B_i}$, $S_{C_i}$, $S_Q$. $Q$ might be reducible, but the groups of states in $S_Q$ *leak* into the $B$ or $C$ blocks, i.e., $S_Q$ represents the *transient* part of the state space. This decomposition is illustrated in Figure 1. We will consider three cases: paths starting from a state in $B_i$, $Q$ or $C_i$. In the first case, for homogeneous and non-homogeneous Markov models (with constant incidence matrix $\tilde{A}_t = \tilde{A}_0$), because $P(x_t \in S_Q | x_{t-1} \in S_Q) < 1$, $\lim_{t \to \infty} P(x_t \in S_Q | x_0 \in S_Q) = 0$. In the second case, because the $B_i$ are primitive, we can apply Theorem 1 to these sub-matrices, and starting from a state in $S_{B_i}$, all information about an initial state at $t_0$ is gradually lost.

### 3.5 Periodic Graphs

A more difficult case to analyze is the third case, i.e., that of paths from state $j$ at time $t_0$ to state $k$ at time $t$, with initial state $j \in S_{C_i}$ associated to a periodic block. Let $d_i$ be the period of the i[th] periodic block $C_i$. It can be shown (Seneta, 1981) that taking $d$ products of periodic matrices with the same incidence matrix and period $d$ yields a block-diagonal matrix whose $d$ blocks are primitive. Thus a product $C^{(t_0,t)}$ retains information about the initial block in which $x_{t_0}$ was. However, for every such block of size $> 1$, information will be gradually lost about the exact identity of the state *within* that block.

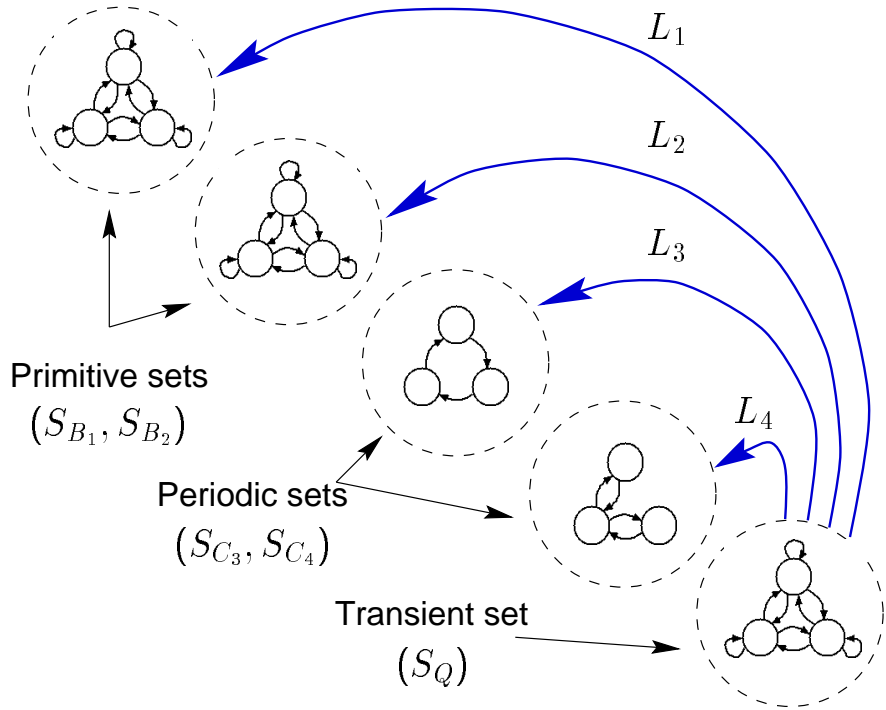

Figure 1: Transition graph corresponding to the canonical decomposition. Large dotted circles represent subgroups of states associated to submatrices $B_i$, $C_i$, and $Q$ in equation (5). The large arrows on the upper right area generically represent transitions from some states in $Q$ to some states in $B_i$ and $C_i$. Transitions among states in each subgroup are depicted inside the large circles.

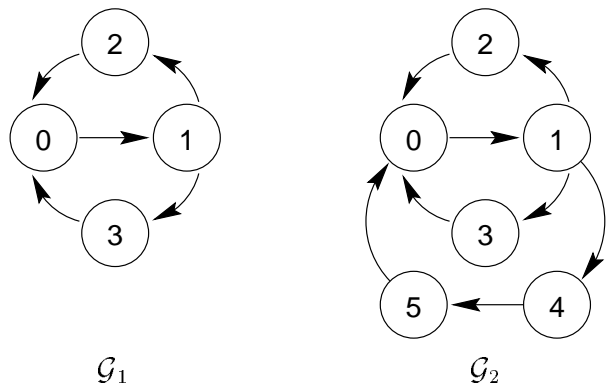

Figure 2: Periodic $\mathcal{G}_1$ becomes primitive (period 1) $\mathcal{G}_2$ when adding loop with states 4,5.

This is best demonstrated through a simple example. Consider the incidence matrix represented by the graph $\mathcal{G}_1$ of Figure 2. It has period 3 and the only non-deterministic transition is from state 1, which can yield into either one of two loops. When many stochastic matrices with this graph are multiplied together, information about *the loop* in which the initial state was is gradually lost (i.e., if the initial state was 2 or 3, this information is gradually lost). What is retained is the *phase* information, i.e., in which block ({0}, {1}, or {2,3}) of a cyclic chain was the initial state. This suggests that it will be easy to learn about the type of outputs associated to each block of a cyclic chain, but it will be hard to learn anything else. Suppose now that the sequences to be modeled are slightly more complicated, requiring an extra loop *of period 4* instead of 3, as in Figure 2. In that case $A$ is primitive: all information about the initial state will be gradually lost.

## 4. Representing and Learning Long-Term Context

Based on the analysis of the previous section, which apply both the homogeneous and non-homogeneous cases, we find in this section that in order to absolutely avoid all diffusion of context and credit information (both *learning* and *representing* context), the transitions should be deterministic (0 or 1 probability). For HMMs, this unfortunately corresponds to a system that can only model cycles (and is therefore not very useful for most applications). Both learning and representing context are hurt by the same ergodicity phenomenon because the state to next state transformation is linear, i.e., forward and backward propagation are symmetrical.

We discuss the practical impact of this ergodicity problem for incremental learning algorithms (such as EM and gradient ascent in likelihood).

### 4.1 Learning Long-Term Dependencies: a Discrete Problem?

To better understand the problem, it is interesting to look at a particular instance of the EM algorithm for HMMs, more specifically, at a form of the update rule for transition probabilities,

$$A_{ij} \leftarrow \frac{A_{ij} \frac{\partial L}{\partial A_{ij}}}{\sum_j A_{ij} \frac{\partial L}{\partial A_{ij}}}. \tag{6}$$

where $L$ is the likelihood of the training sequences. We might wonder if, starting from a positive stochastic matrix, the learning algorithm could learn the topology, i.e., replace some transition probabilities by zeroes. Starting from $A_{ij} > 0$ we could obtain a new $A_{ij} = 0$ only if $\frac{\partial L}{\partial A_{ij}} = 0$, i.e., on a local maximum of the likelihood $L$. Thus the EM training algorithm will not *exactly* obtain zero probabilities. Transition probabilities might however *approach* 0. Furthermore, once $A_{ij}$ has taken a near-zero value, it will tend to remain small. This suggests that prior knowledge (or initial values of the parameters), rather than learning, should be used, if possible, to determine the important elements of the topology, and for establishing the long-term relations between elements of the observed sequences.

It is also interesting to ask in which conditions we are guaranteed that there will not be any diffusion (of influence in the forward phase, and credit in the backward phase of training). It requires that all of the eigenvalues have a norm that is 1. This can be achieved with periodic matrices $C$ (of period $d$), which have $d$ eigenvalues that are the $d$ roots of 1 on the complex unit circle. To avoid any loss of information also requires that $C^d = I$ be the identity, since any diagonal block of $C^d$ with size more than 1 will bring a loss of information (because of ergodicity of primitive matrices). This can be generalized to reducible matrices whose canonical form is composed of periodic blocks $C_i$ with $C_i^d = I$.

The condition we are describing actually corresponds to a matrix with only 1's and 0's. For this type of matrix, the incidence matrix $\tilde{A}_t$ of $A_t$ is equal to the matrix $A_t$ itself. Therefore, when $\tilde{A}_t$ is fixed, the Markov chain is also homogeneous. It appears that many interesting computations cannot be achieved with such constraints (i.e., only allowing one or more cycles of the same period and a purely deterministic and homogeneous Markov chain). Furthermore, if the parameters of the system are the transition probabilities themselves (as in ordinary HMMs), such solutions correspond to a subset of the corners of the 0-1 hypercube in parameter space. Away from those solutions, learning is mostly influenced by *short term* dependencies, because of diffusion of credit. Furthermore, as seen in equation (6), algorithms like EM will tend to stay near a corner once it is approached. This suggests that *discrete optimization* algorithms, rather continuous local algorithms, may be more appropriate to explore the (legal) corners of this hypercube.

Examples of to this approach are found in the area of grammar inference for natural language modeling (e.g., variable memory length Markov models, Ron et al., 1994, or constructive algorithms for learning context-free grammars, Lari & Young, 1990, Stolcke & Omohundro, 1993). The problem of diffusion studied here applies only to algorithms that use gradient information (such as the Baum-Welch and gradient-based algorithms) and a gradual modification of transition probabilities. It would be interesting to evaluate how such constructive and discrete search algorithms perform when properly solving the task requires to learn to represent long-term context. On the basis of the results of this paper, however, we believe that in order to successfully learn long-term dependencies, such algorithms should look for very sparse topologies (or very deterministic models). Note that some of the already proposed approaches (Ron et al., 1994) are limited in the type of context that can be represented (e.g., no loops in the graph and the constraint that all intermediate observations between times $t_0$ and $t$ must be represented by the state variable in order to model the influence of $\boldsymbol{y}_{t_0}$ on $\boldsymbol{y}_t$).

### 4.2 Diffusion of Credit

We have already found above that except in the special case of 0 or 1 transition probabilities, the state variable becomes more and more independent of remote past states (and therefore of remote past inputs and outputs). Since this prevents robustly representing long-term context, learning such a long-term context is also made more and more difficult for longer term dependencies.

Diffusion in Markovian ModelsHowever, it is interesting to consider how the ergodicity of the transition probability matrix directly affects the the forward-backward equations (2) (to propagate context information forward and backward) used in learning algorithms such as EM and (implicitly) gradient descent. In particular, let us consider the Dobrushin ergodicity coefficient of these matrix products. First, let $V_t = A_t \Lambda_t \cdots A_T \Lambda_T$, then

$$\tau_1(V_t) \leq \tau_1(A_t)\tau_1(\Lambda_t) \cdots \tau_1(A_T)\tau_1(\Lambda_T) = \prod_{\tau=t}^{T} \tau_1(\Lambda_\tau)\tau_1(A_\tau) \qquad (7)$$

We have already seen that $\tau_1(A_t) < 1$ unless the transition probabilities are all 0 or 1. Remember that the emission probability matrix $\Lambda_\tau$ is diagonal. Applying the definition of $\tau_1$ (equation 4) to a diagonal matrix $D$, we obtain

$$\tau_1(D) = \frac{1}{2} \max_{i,j} \Big( |D_{ii} - D_{ij}| + |D_{ij} - D_{jj}| \Big) = \frac{1}{2} \max_{i,j} \Big( |D_{ii}| + |D_{jj}| \Big) \text{ with } i \neq j.$$

Therefore,

$$\tau_1(\Lambda_t) = \frac{1}{2} \max_{i,j} \Big( P(\boldsymbol{y}_t|x_t = i, \boldsymbol{u}_t) + P(\boldsymbol{y}_t|x_t = j, \boldsymbol{u}_t) \Big) \text{ with } i \neq j,$$

which is the average of the two largest emission probabilities at this time step. Therefore, when the transition probabilities are not all 0 or 1, in the case of discrete outputs, $\tau_1(\Lambda_t) \leq 1$, and the ergodicity coefficient of the matrix product $V_t$ in equation (7) converges to 0 as $T - t$ increases. Note that this product gives the gradient of $\alpha_{i,T}$ with respect to $\alpha_{j,t}$ (from equation 1) and is used in the EM algorithm (Baum et al., 1970; Levinson et al., 1983) as well as in gradient-based algorithms (Bridle, 1990; Bengio, De Mori, Flammia, & Kompe, 1992; Bengio & Frasconi, 1995b).

For example, in the case of a learning criterion $L$,

$$\frac{\partial L}{\partial \alpha_t} = V_t \frac{\partial L}{\partial \alpha_T}$$

where $\frac{\partial L}{\partial \alpha_t}$ is the vector $[\frac{\partial L}{\partial \alpha_{1,t}} \ldots \frac{\partial L}{\partial \alpha_{n,t}}]$. Since $V_t$ is used to propagate credit backwards, its convergence to rank 1 means that long-term credit is gradually lost as it is propagated backwards: the gradient of the learning criterion with respect to all the past states becomes the same, i.e., $\frac{\partial L}{\partial \alpha_t}$ converges to a multiple of $[1, 1, \ldots, 1]$.

The continuous emissions case is more difficult because the density $P(\boldsymbol{y}_t|x_t = i, \boldsymbol{u}_t)$ can locally be greater than one. The above result can still be obtained if we restrict our attention to the cases in which the product of the largest emission probabilities at each time step is bounded, which is the most likely in practice. In the case where it is not bounded, we conjecture that the same result can be obtained by considering scaled emission probability matrices, with a scaling factor $s_t$ that is 1 when the emission probability is less than 1, and that is $1/\max_i P(\boldsymbol{y}_t|x_t = i, \boldsymbol{u}_t)$ otherwise. Letting $U_t = A_t s_t \Lambda_t \cdots A_T s_T \Lambda_T$, although the overall gradient with respect to all the past states can grow very large (as $T - t$ increases), the rank of $U_t$ still converges to 1, and the vector $\boldsymbol{\beta}_t = \frac{\partial L}{\partial \boldsymbol{\alpha}_t}$ also converges to a (possibly very large) multiple of $[1, 1, \ldots, 1]$.

261

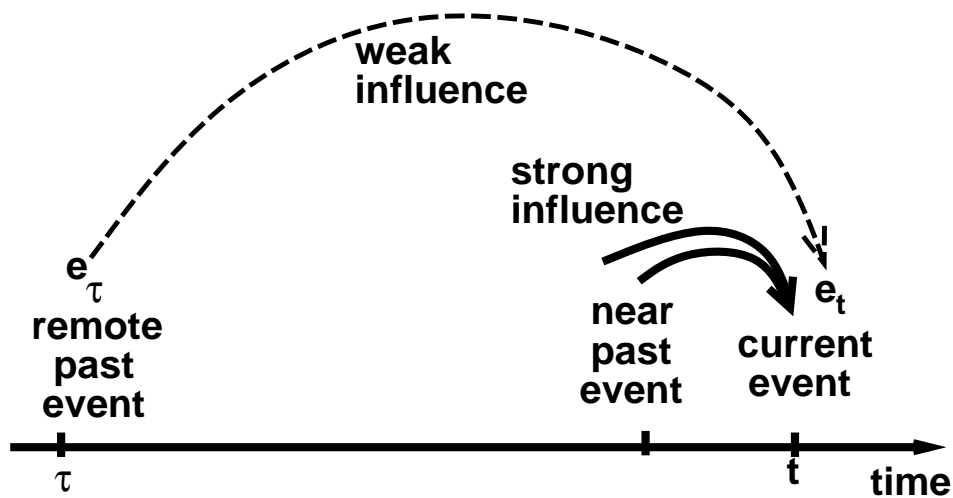

Figure 3: Typical problem with short-term dependencies hiding the long-term dependencies.

In practice we train HMMs with finite sequences. However, training will become more and more numerically ill-conditioned as one considers longer term dependencies. Consider as in Figure 3 two events $e_t$ (occurring at time $t$) and $e_\tau$ (occurring at time $\tau$ much earlier than $t$), and suppose there are also "interesting" events occurring in between (i.e., events which should influence the state variable at time $t$ in order to better model outputs at time $t$ or later). Let us consider the overall influence of states at times $s < t$ upon the likelihood of the outputs at time $t$. Because of the phenomenon of diffusion of credit, and because gradients are added together, the influence of intervening events (especially those occurring shortly before $t$) will be much stronger than the influence of $e_\tau$. Furthermore, this problem gets *geometrically* worse as $t - \tau$ increases.

### 4.3 Sparse Matrices and Prior Knowledge

Clearly a positive matrix (corresponding to a fully-connected graph) is primitive. Thus in order to learn long-term dependencies, we would like to have many zeros in the matrix of transition probabilities (which reduces the problem of diffusion, as confirmed by the experiments described in Section 5 and illustrated in Figure 5). Unfortunately, this generally supposes *prior knowledge* of an appropriate connectivity graph. In practical applications of HMMs, for example to speech recognition (Lee, 1989; Rabiner, 1989) or protein secondary structure modeling (Chauvin & Baldi, 1995), prior knowledge is heavily used in setting up the connectivity graph. As illustrated in Figure 4, in speech recognition systems the meaning of individual states is usually fixed a-priori except within phoneme models. The representation of long-term context is therefore not learned by the HMM. Transition probabilities between groups of states representing a phoneme in a certain context are "learned" from text or labeled speech data. However, in that case the "model" is a Markov model, not a *hidden* Markov model: learning consists in counting co-occurrence of events such as phonemes or words. The *hard problem* of learning a representation of context is therefore avoided by choosing it on the basis of prior knowledge.

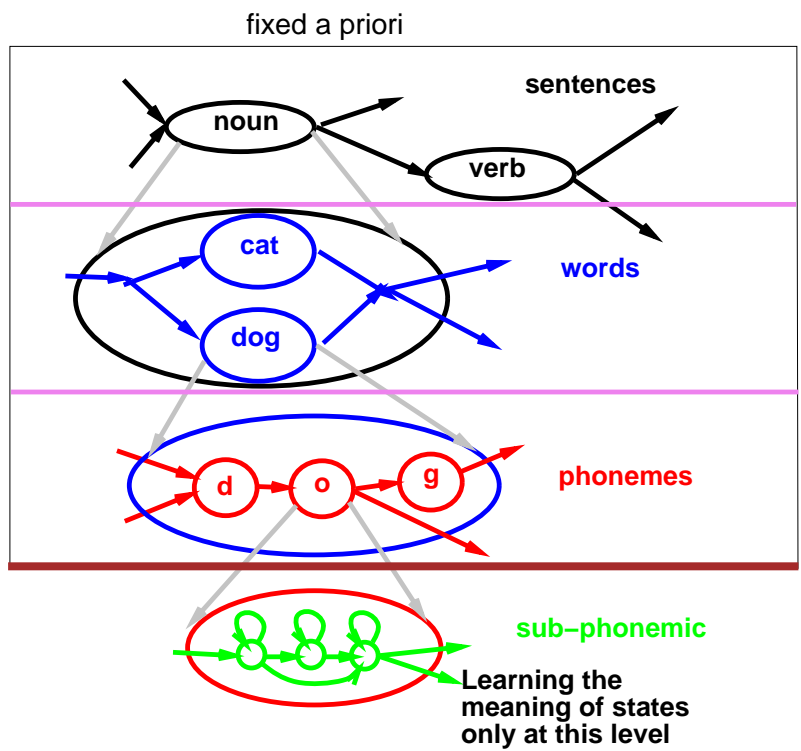

Figure 4: Learning of a representation of context in speech recognition HMMs is typically limited to what happens within a phoneme. Higher-level representations are chosen from prior knowledge and those parameters are often estimated from simple co-occurrence statistics.

Another direction of research should be in ways to incorporate some prior knowledge with learning from examples, preferably in a way that simplifies the problem of learning (new) long-term dependencies. Our current research in this direction is based on the old AI idea of using a multi-scale representation. The state variable is decomposed into several "sub-state" variables (whose Cartesian product is equal to the "full" state variable), each operating at a different time scale. The a-priori assumption is that long-term context will be represented by "slow" state variables, which must be insensitive to the precise timing of events. This allows the propagation of context (and credit, for learning) over long durations through those higher-level state variables. To impose these multiple time scales, one can introduce constraints on the transition probabilities, such that the "slow" variables always have a small probability of changing at any time step. Another useful assumption is that the transition probabilities can be factored in terms of the conditional sub-state probabilities at each time scale, given the full state. We conjecture that the hypothesis behind this multi-scale structure is appropriate for most "natural" sequence learning tasks (such as those humans perform).

## 5. Experiments

In this section we report some experimental results. Firstly, we study, from a numerical point of view, the convergence of products of stochastic matrices. Then we report an example of training in a problem in which the span of temporal dependencies is artificially controlled.

### 5.1 Diffusion: Numerical Simulations

In this experiment we measure how (and if) different kinds of products of stochastic matrices converged, for example to a matrix with equal rows. We ran 4 simulations, each with an 8-state non-homogeneous Markov chain but with different constraints on the transition graph: 1) $\mathcal{G}$ is fully connected; 2) $\mathcal{G}$ is a left-to-right model (i.e., the incidence matrix $\tilde{A}$ is upper triangular); 3) $\mathcal{G}$ is left-to-right but only one-state skips are allowed (i.e., $\tilde{A}$ is upper bidiagonal); 4) $A_t$ are periodic with period 4. Results shown in Figures 5 and 6 confirm the convergence towards zero of the ergodicity coefficient at a rate that depends on the graph topology. The exception is, as expected, the case of periodic matrices. Note how the sparser graphs have a larger ergodicity coefficient, which should ease the learning of long-term dependencies. In Figure 6, we represent visually the convergence of fully connected matrices to row equality, in only 4 time steps, towards equal rows. Each of the transition probability matrices $A_t$ ($t = 1, 2, 3, 4$) was chosen randomly from a uniform distribution.

### 5.2 Training Experiments

To evaluate how diffusion impairs training, a set of controlled experiments were performed, in which the training sequences were generated by a simple homogeneous HMM with long-term dependencies, depicted in Figure 7.

Two branches generate similar sequences except for the first and last symbol. The extent of the long-term context is controlled by the self transition probabilities of states 2 and 5, $\lambda = P(x_t = 2|x_{t-1} = 2) = P(x_t = 5|x_{t-1} = 5)$. Span or "half-life" is $\log(.5)/\log(\lambda)$, i.e., $\lambda^{\text{span}} = .5$). Following Bengio et al. (1994), data was generated for various span of long-term dependencies (0.1 to 1000).

For each series of experiments, varying the span, 20 different training trials were run per span value, with 100 training sequences[3]. Training was stopped either after a maximum number of epochs (200), of after the likelihood did not improve significantly, i.e., $(l(t) - l(t-1))/|l(t)| < 10^{-5}$, where $l(t)$ is the logarithm of the likelihood of the training set at epoch $t$. A trial is considered successful (converged) when it yields a likelihood almost as good or better than the likelihood of the generating HMM on the same data.

If the HMM is fully connected (except for the final absorbing state) and has just the right number of states, trials *almost never converge* to a good solution (1 in 160 did). Increasing the number of states and randomly putting zeroes in the transition matrix helps convergence. This confirms common intuition, although using more states than strictly necessary may result in worse generalization to new examples and, hence, may not be an advisable solution to solve convergence problems. The randomly connected HMMs had 3

---

3. This relatively small number of training sequences appeared sufficient since the likelihood of the generating HMM did not improve much when trained on this data.

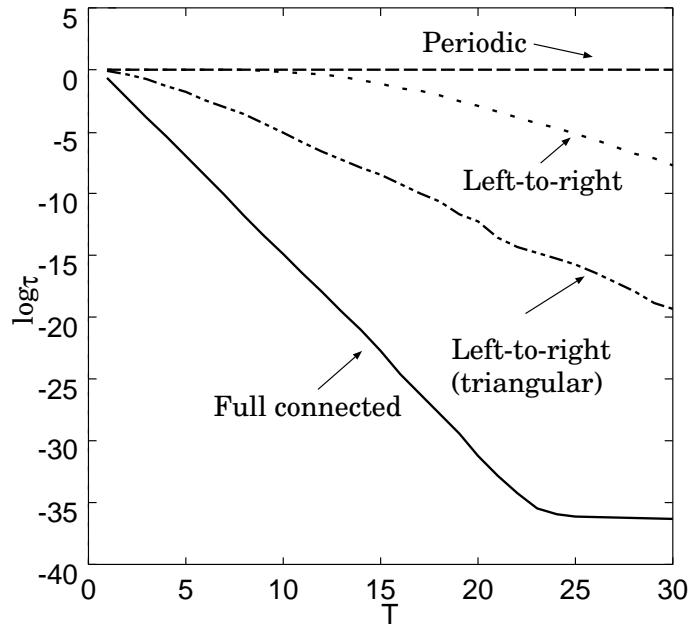

Figure 5: Convergence of Dobrushin's coefficient (see Definition 6) in product of stochastic matrices associated to non-homogeneous Markov chains constrained by different transition graphs. The flattening of the bottom curve is due to the limits of numerical precision in the computer experiments.

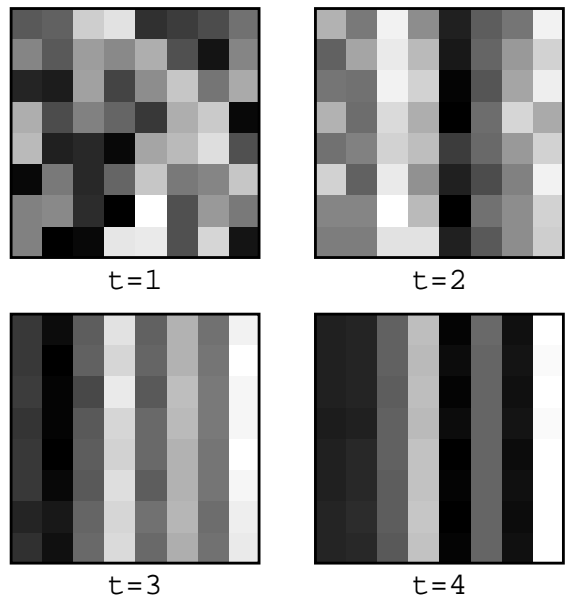

Figure 6: Evolution of matrix products $A^{(1,t)}$ for a model having a fully connected transition graph. Matrix elements (transition probabilities) are visualized with gray levels.

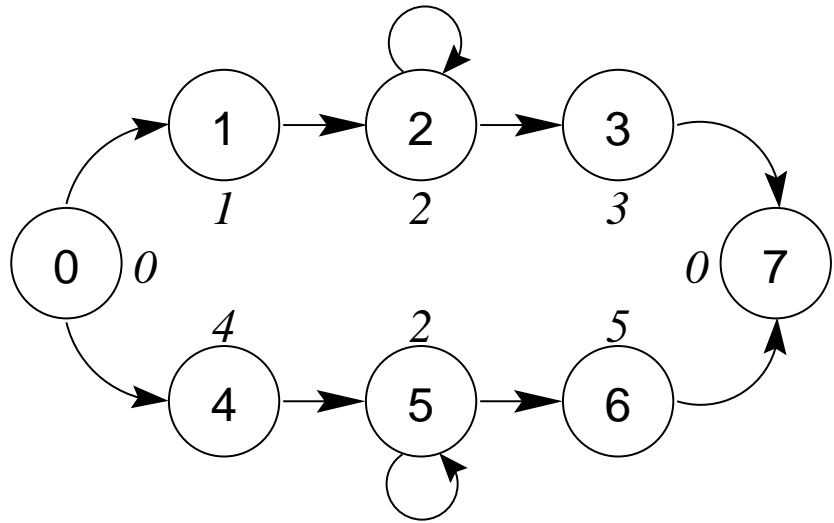

Figure 7: Generating HMM. Numbers in state circles denote state indices, numbers out of state circles denote output symbols. This HMM was used to generate the training data for the experiments summarized in Figure 8.

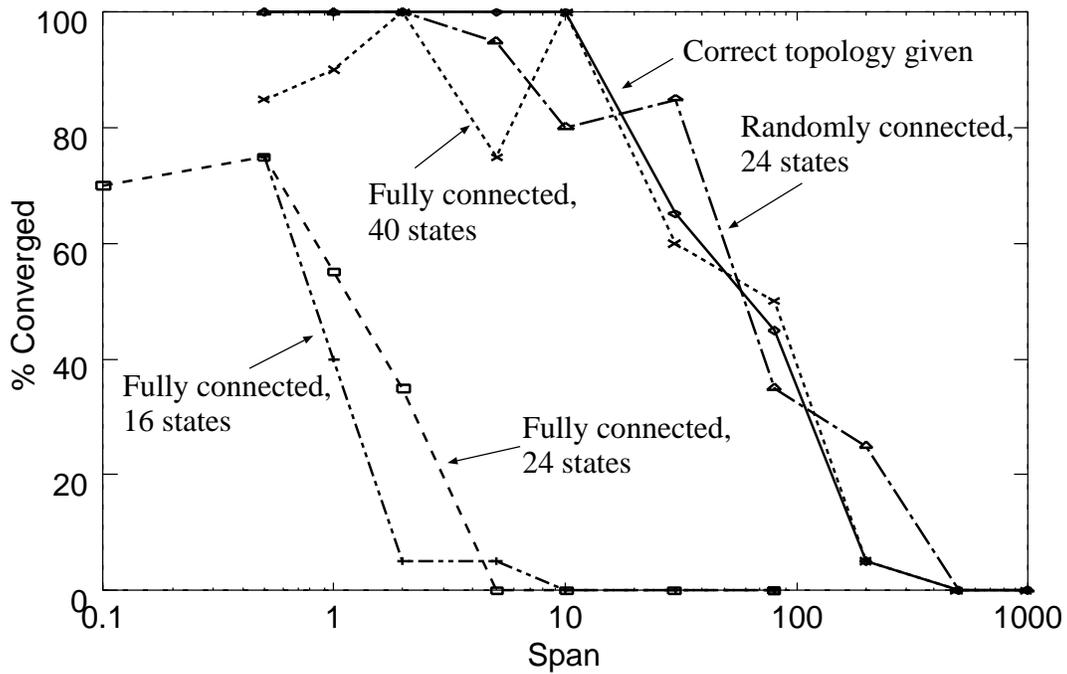

Figure 8: Percentage of convergence to a good solution (over 20 trials) for various series of experiments as the span of dependencies is increased (by increasing the self-transition probabilities of states 2 and 5). The task consists in modeling sequences generated by the HMM depicted in Figure 7.

Diffusion in Markovian Modelstimes more states than the generating HMM and random connections were created with 20% probability. Figure 8 shows the average number of converged trials for these different types of HMM topologies. In all cases the number of successful trials rapidly drops to zero beyond some value of span. In failed trials, the equivalent of states 3 and 6 of the generating HMM are usually confused, i.e., these solutions don't take the beginning of the sequence into account to represent the distribution of the symbols near the end of the sequence. It is interesting to note that HMMs with many more states than necessary but sparse connectivity performed much better. Typically, a sparser graph corresponds to a larger coefficient of ergodicity (as exemplified in Figure 5), which allows long-term dependencies to be represented and learned more easily.

Another interesting observation is that in many cases, the training curve goes through one or more very flat plateaus. Such plateaus could be explained by the diffusion problem: the relative gradient with respect to some parameters is very small (thus the algorithm appears to be stuck). These plateaus can become a very serious problem when their slope approaches numerical precision or their length becomes unacceptable.

## 6. Conclusion and Future Work

In previous work on recurrent networks (Bengio et al., 1994) we had found that, for these nonlinear dynamical parameterized systems, propagating credit over the long term was incompatible with storing information for the long term. Basically, with enough non-linearity (larger weights) to store long-term context robustly, gradients back-propagated through time vanish rapidly. In this paper, we have also found negative results concerning the representation and learning of long-term context, but they apply to Markovian models such as HMMs, IOHMMs or POMDPs. For these models, we found that both the representation and the learning of long-term context information are tied together. In general, they are both hurt by the ergodicity of the transition probability matrix (or submatrices of it). However, when the transition probabilities are close to 1 and 0, information can be stored for the long term *and* credit can be propagated over the long term. Like our findings for recurrent networks, this suggests that the problem of learning long-term dependencies looks more like a *discrete* optimization problem. It appears difficult for local learning algorithm such as EM or gradient descent to learn optimal transition probabilities near 1 or 0, i.e., to learn the topology, while taking into account long-term dependencies. This should encourage research on alternative (discrete) algorithms for discovering HMM topology (especially for representing long-term context), such as those proposed by Stolcke & Omohundro (1993) and Ron et al. (1994). Our results suggest that such algorithms should strive to discover sparse topologies, or almost deterministic models. The arguments presented here are essentially an application of established mathematical results on Markov chains to the problem of learning long term dependencies in homogeneous and non-homogeneous HMMs. These arguments were also supported by experiments on artificial data, studying the phenomenon of diffusion of credit and the corresponding difficulty in training HMMs to learn long-term dependencies.

IOHMMs (Bengio & Frasconi, 1994, 1995b) and POMDPs (Sondik, 1973, 1978; Chrisman, 1992) are non-homogeneous variants of HMMs, i.e., the transition probabilities are function of the input (for IOHMMs) or the action (for POMDPs) at each time $t$. The re-

267

sults of this paper suggests that such non-homogeneous Markovian models could be better suited (in some situations) to representing and learning long-term context. For such models, forcing transition probabilities to be near 0 or 1 still allows the system to model some interesting phenomena and perform useful computations. In practice, this means that the underlying dynamics of state evolution to be modeled should be deterministic. For example, a deterministic IOHMM can recognize strings from a deterministic grammar, taking into account long-term dependencies (Bengio & Frasconi, 1995b). For HMMs this constraint restricts the model to simple cycles, which are not very interesting.

Our analysis and numerical experiments also suggest that using many more hidden states than necessary, with a sparse connectivity, reduces the diffusion problem. Another related issue to be investigated is whether techniques of symbolic prior knowledge injection (see, e.g., Frasconi, Gori, Maggini, & Soda, 1995) can be exploited to choose good topologies, or combine specific a-priori knowledge with learning from examples.

Based on the analysis presented here, we are also exploring another approach to learning long-term dependencies that consists in building a hierarchical representation of the state. This can be achieved by introducing several sub-state variables whose Cartesian product corresponds to the system state. Each of these sub-state variables can operate at a different time scale, thus allowing credit to propagate over long temporal spans for some of these variables.

## Acknowledgments

Yoshua Bengio is also with the adaptive systems department at AT&T Bell Labs (Holmdel, NJ). We would like to thank Léon Bottou for his many useful comments and suggestions, and the NSERC, FCAR, and IRIS Canadian funding agencies for support.